\documentclass{article}

\usepackage{microtype}
\usepackage{graphicx}
\usepackage{subfigure}
\usepackage{booktabs} 

\usepackage[table]{xcolor}
\usepackage{arydshln}
\usepackage{multirow}
\usepackage{multicol}
\usepackage{amsfonts}
\usepackage{threeparttable}

\usepackage{hyperref}

\usepackage[ruled]{algorithm2e}

\usepackage[accepted]{icml2025}

\usepackage{amsmath}
\usepackage{amssymb}
\usepackage{mathtools}
\usepackage{amsthm}

\usepackage[capitalize,noabbrev]{cleveref}

\theoremstyle{plain}

\theoremstyle{definition}

\theoremstyle{remark}

\usepackage[textsize=tiny]{todonotes}

\icmltitlerunning{FGFP: A Fractional Gaussian Filter and Pruning for Deep Neural Networks Compression}

\begin{document}

\twocolumn[
\icmltitle{FGFP: A Fractional Gaussian Filter and Pruning for Deep Neural Networks Compression}



\icmlsetsymbol{equal}{*}

\begin{icmlauthorlist}
\icmlauthor{Kuan-Ting Tu}{equal,GSAT}
\icmlauthor{Po-Hsien Yu}{equal,GIEE}
\icmlauthor{Yu-Syuan Tseng}{GIEE}
\icmlauthor{Shao-Yi Chien}{GIEE}
\end{icmlauthorlist}

\icmlaffiliation{GSAT}{Graduate School of Advanced Technology, National Taiwan University, Taipei, Taiwan}
\icmlaffiliation{GIEE}{Graduate Institute of Electronics Engineering, National Taiwan University, Taipei, Taiwan}

\icmlcorrespondingauthor{Kuan-Ting Tu}{kttu@media.ee.ntu.edu.tw}
\icmlcorrespondingauthor{Po-Hsien Yu}{michaelyu@media.ee.ntu.edu.tw}
\icmlcorrespondingauthor{Yu-Syuan Tseng}{ystseng@media.ee.ntu.edu.tw}
\icmlcorrespondingauthor{Shao-Yi Chien}{sychien@ntu.edu.tw}

\icmlkeywords{Machine Learning, ICML}

\vskip 0.3in
]



\printAffiliationsAndNotice{\icmlEqualContribution} 

\begin{abstract}

Network compression techniques have become increasingly important in recent years because the loads of Deep Neural Networks (DNNs) are heavy for edge devices in real-world applications. While many methods compress neural network parameters, deploying these models on edge devices remains challenging. To address this, we propose the fractional Gaussian filter and pruning (FGFP) framework, which integrates fractional-order differential calculus and the Gaussian function to construct fractional Gaussian filters (FGFs). To reduce the computational complexity of fractional-order differential operations, we introduce Gr\"unwald-Letnikov fractional derivatives to approximate the fractional-order differential equation. The number of parameters for each kernel in FGF is minimized to only seven. Beyond the architecture of Fractional Gaussian Filters, our FGFP framework also incorporates Adaptive Unstructured Pruning (AUP) to achieve higher compression ratios. Experiments on various architectures and benchmarks show that our FGFP framework outperforms recent methods in accuracy and compression. On CIFAR-10, ResNet-20 achieves only a 1.52\% drop in accuracy while reducing the model size by 85.2\%. On ImageNet2012, ResNet-50 achieves only a 1.63\% drop in accuracy while reducing the model size by 69.1\%.

\end{abstract}    
\section{Introduction}
\label{sec:intro}

Over the past decade, computer vision has experienced rapid advancements, primarily driven by the emergence of deep neural networks (DNNs). Modern deep learning models now achieve state-of-the-art performance across various tasks, including image classification, object detection, and pose estimation. To improve accuracy, many tasks leverage an increased number of trainable parameters by expanding the depth or width of models. Although larger models can deliver higher precision, they also have significant drawbacks, such as higher memory storage and access costs. This poses substantial challenges for deploying models on edge devices in real-world applications, such as smartphones, laptops, or resource-constrained embedded systems.

Many studies have explored efficient techniques to deploy large models on edge devices. Unstructured pruning \cite{Zhang_systematic, Han_Learning, Frankle_Lottery} preserves accuracy, but remains high complexity of computing. Structured pruning \cite{Tang_SCOP, Yuan_Growing, Zhou_Efficient, Wang_All-in-One, Yuan_ARPruning, Van_Efficient, Liu_Soft} reduces computation by removing kernels or channels, though it often sacrifices important features and leads to significant accuracy loss. Low-rank compression \cite{Denton_Exploiting, Phan_Stable, Chu_Low-rank, Li_Heuristic, Guo_Compact, Liu_TDLC, Sui_ELRT} provides another solution by decomposing the kernels into lower-rank forms, reducing both parameters and computation. Recent hybrid methods \cite{Li_Hinge, Li_Towards, Ruan_EDP, Pham_Enhanced} combine pruning and low-rank techniques to achieve higher compression with low accuracy degradation.

The motivation for FGFP comes from previous work \cite{Zamora_Convolutional}, which successfully integrated the traditional computer vision filters, such as the Gaussian, Laplacian \cite{Gaussian_Laplacian}, and Sobel \cite{Sobel} filter with CNNs through fractional-order differentiation. FGFP shares the same fractional Gaussian filter (FGF) parameters across all channels in each kernel to reduce parameters. However, applying the same filter can hurt accuracy, so we apply the channel-attention mechanism to minimize the accuracy degradation and design two types of FGFs: the channel-attention fractional Gaussian filter (CA-FGF) and three-dimensional fractional Gaussian filter (3D-FGF). Furthermore, we leverage the Gr\"unwald–Letnikov fractional derivative \cite{Jalalinejad_simple} to approximate fractional differential equations to simplify the computational complexity of fractional-order differential operations. To maximize the compression ratio, we also explore incorporating the adaptive unstructured pruning (AUP) strategy into our FGFP.

In summary, the main contributions of this work are as follows:
\begin{itemize}
 \item We propose the fractional Gaussian filter and pruning (FGFP) framework, which combines the fractional Gaussian filter (FGF) and adaptive unstructured pruning (AUP) to reduce the number of parameters significantly.
 \item We use the channel-attention mechanism to design two forms of the fractional Gaussian filter (FGF): the channel-attention fractional Gaussian filter (CA-FGF) and the three-dimensional fractional Gaussian filter (3D-FGF). Both the FGFs perform excellently in removing the redundant parameters with low degradation of model accuracy. Moreover, we introduce the Gr\"unwald–Letnikov fractional derivative into the FGF and simplify the fractional derivative to the trinomial polynomial.

 \item We conducted comprehensive experiments with state-of-the-art methods on two benchmarks, CIFAR-10 and ImageNet2012. The experimental results illustrate our FGFP's effectiveness, and the FGFP outperforms several recent works. For instance, when evaluating the FGFP with ResNet-20 on CIFAR-10, our method achieves only a 1.58\% drop in accuracy while reducing the model size by 85.1\%. Also, on ImageNet2012, our method achieves only a 1.63\% drop in accuracy while reducing the model size by 69.1\% with ResNet-50.
 
\end{itemize}

\section{Proposed Methods}

\begin{figure*}[htb]
    \centering
    \includegraphics[width=0.99\linewidth]{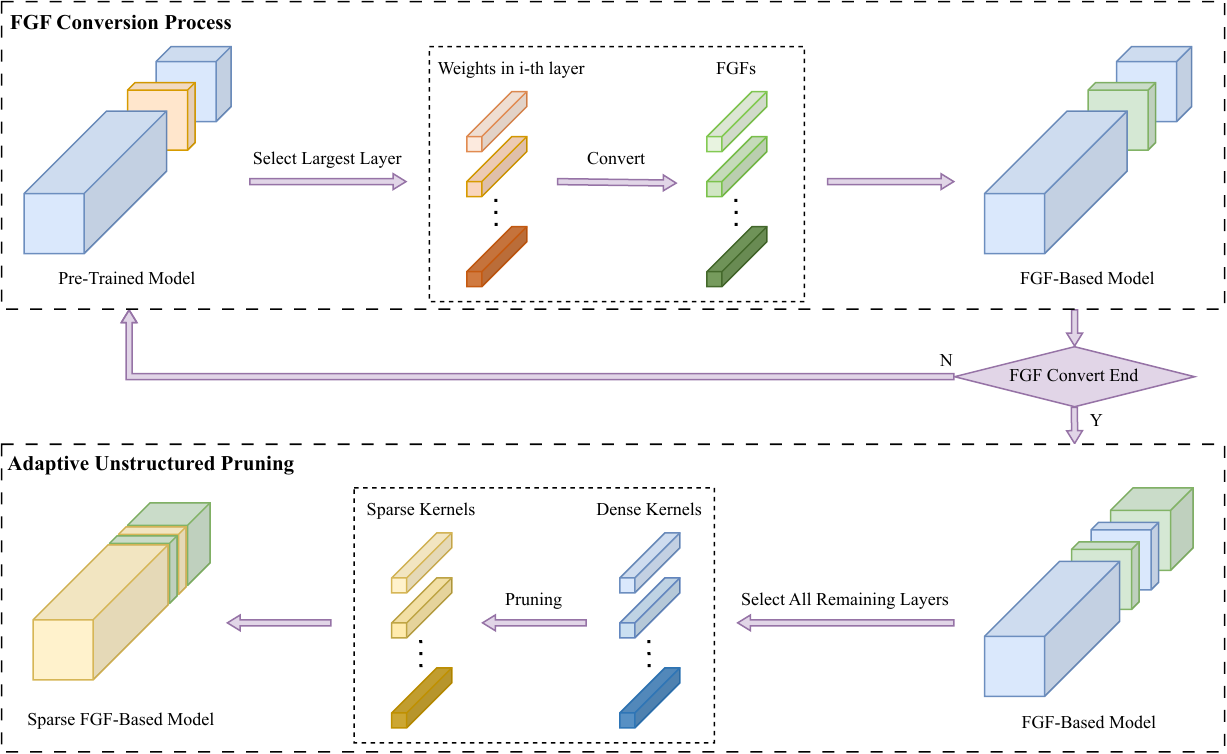}
    \caption{
    Overview of the FGFS methodology. The FGF conversion process begins with selecting the largest layer in the pre-trained model and then converting weights into FGF representations to generate the FGF-based model. This process repeats until all selected layers are converted. The AUP process, where the remaining layers of the FGF-based model undergo adaptive unstructured pruning, transforms dense kernels into sparse kernels. The final result is a sparse FGF-based model optimized for performance and efficiency.
    }
    \label{fig:Overall_HFGFS_Method}
\end{figure*}

In this section, we will introduce the detailed Fractional Gaussian Filter and Pruning (FGFP), and the overall methods are shown as Fig.~\ref{fig:Overall_HFGFS_Method}.

\subsection{Gr\"unwald-Letnikov Fractional Derivatives}
Ordinarily, the definition of $n$th-order derivative of function $f$ is \cite{Jalalinejad_simple}:
\begin{equation} \label{diff_basic_definitiion}   
f^{(n)}(x) = \frac{d^nf}{dx^n} = \lim_{h \rightarrow 0}\frac{1}{h^n}\sum^n_{r=0} (-1)^r 
\begin{pmatrix}
  n \\ 
  r
\end{pmatrix} f(x - rh).
\end{equation}
According to the Gr\"unwald-Letnikov fractional derivatives for univariate function $f$ which mentioned in \cite{Gldefinition, Jalalinejad_simple}, we can redefined eq.~(\ref{diff_basic_definitiion}) as belows:
\begin{equation} \label{limgl}
D^{\alpha}_{G-L} f(x) = \lim_{h \rightarrow 0} \frac{1}{h^\alpha}\sum^{\left [\frac{x-a}{h}\right ]}_{r=0} (-1)^r 
\begin{pmatrix}
    \alpha \\
    r
\end{pmatrix}f(x - rh),
\end{equation}
where 
$$\begin{pmatrix}
    \alpha \\
    r
\end{pmatrix} = \frac{\Gamma (\alpha + 1)}{\Gamma (r+1) \Gamma (\alpha - r + 1) }$$
and $\Gamma$ is the gamma function. We can also approximate and expand eq.~(\ref{limgl}) as follows:
\small
\begin{equation} 
\begin{aligned}
D^\alpha_{G-L} f(x) &\approx f(x) + (-\alpha)f(x-1) + \frac{(-\alpha)(-\alpha+1)}{2}f(x-2) \\
&+ \cdots + \frac{\Gamma (-\alpha+ 1)f(x-n)}{n!\Gamma (-\alpha + n + 1)}.
\end{aligned}
\end{equation}
\small
\begin{equation} 
\label{eq:gl_single}
\begin{aligned}
D^\alpha_{G-L} f(x) &\approx f(x) - \alpha f(x-1) + \frac{\alpha(\alpha - 1)}{2}f(x-2) \\
&+ \epsilon^\alpha_x f(x),
\end{aligned}
\end{equation}
where $\epsilon^\alpha_x f(x)$ is the corresponding approximate error, and the error can be ignored in eq.~(\ref{eq:gl_single}). Thus, the Gr\"uwald-Letnikov derivative can be simplified as follows:
\begin{equation} 
\label{eq:gl_approximate}
D^\alpha_{G-L} f(x) \approx f(x) -\alpha f(x-1) + \frac{\alpha(\alpha - 1)}{2}f(x-2),
\end{equation}

\subsection{Fractional Gaussian Filter (FGF)}
The Gaussian filter is widely used in image processing and signal processing, primarily to reduce high-frequency noise and detail in images or signals while preserving crucial structural information. The Gaussian filter applies a convolution operation using a kernel derived from the Gaussian function, and the 2D Gaussian function is defined as:
\begin{equation} \label{eq:2Dgaussian}
G(x,y) = e^{-\frac{(x - x_0)^2 + (y - y_0)^2}{\sigma^2}},
\end{equation}
where $\sigma$ represents the standard deviation of the Gaussian distribution and $x_0$ and $y_0$ are the positions of the center of the peak. Notably, the 2D Gaussian function can be decomposed into the product of two 1D Gaussian filters, and we can redefined eq.~(\ref{eq:2Dgaussian}) as below:
\begin{equation} \label{eq:2D_gaussian_decompose}
\begin{aligned}
G(x,y) &= e^{-\frac{(x - x_0)^2 + (y - y_0)^2}{\sigma^2}} \\
&= e^{-\frac{(x - x_0)^2}{\sigma^2}} \cdot e^{-\frac{(y - y_0)^2}{\sigma^2}} = G(x)G(y),
\end{aligned}
\end{equation}
where $G(x)$ and $G(y)$ are the 1D Gaussian function in x- and y- direction.
Furthermore, the first and second derivatives of the Gaussian filter are the Sobel filter and the Laplacian filter, commonly used to detect edges in images by removing the low-frequency features. Since different order derivatives of the Gaussian filter can extract features from various frequencies, general frequency-adjustable filters can be obtained by applying the fractional derivatives to the Gaussian filter \cite{Zamora_Convolutional}. Hence, the fractional Gaussian filters (FGF), can be defined as follows:
\begin{equation} \label{eq:fractional_gaussian_filter}
\begin{aligned}
F_{fg} &= D^a_x D^b_y G(x,y) = D^a_x D^b_y G(x)G(y) \\
       &= D^a_x G(x) \times D^b_y G(y),
\end{aligned}
\end{equation}
where $F_{fg}$ denotes the fractional Gaussian filters, $D$ is the fractional derivatives, $a$ and $b$ are the order of the fractional derivatives. Moreover, to simplify the computation of the fractional derivatives, we introduce the Gr\"unwald-Letnikov Fractional Derivatives. According to eq.~(\ref{eq:gl_approximate}), we can simplify $D^a_x G(x)$ and $D^b_y G(y)$ as below:
\begin{equation} \label{eq:fodiffx} 
    D^a_x G(x) = G(x) - a G(x-1) + \frac{a(a-1)}{2}G(x-2);
\end{equation}
\begin{equation} \label{eq:fodiffy}
    D^b_y G(y) = G(y) - b G(y-1) + \frac{b(b-1)}{2}G(y-2).
\end{equation}

In the original fractional Gaussian filter, there are five parameters for each channel, $a$, $b$, $x_0$, $y_0$, and $\sigma$. Also, the range of them would be limited as follows: $a \in [0,2]$, $b \in [0,2]$, $x_0 \in (-\infty,\infty)$, $y_0 \in (-\infty, \infty)$, and $\sigma \in (0,\infty)$. The fractional orders $a$ and $b$ are constrained to the range $[0, 2]$ to ensure that the learned functions remain within the domain of traditional filters, facilitating the identification of suitable filter forms.
The remaining parameters follow the common settings of the classical Gaussian function. However, generating the FGF for each channel results in poor model compression efficiency. Specifically, in the case of a $3\times 3$ filter, the number of trainable parameters is reduced from nine to five, achieving only a compression ratio of $44.4\%$. To address this problem, we propose two types of FGF: the channel-attention fractional Gaussian filter (CA-FGF) and the three-dimensional fractional Gaussian filter (3D-FGF).

\begin{figure*}[htb]
    \centering
    \includegraphics[width=0.95\linewidth,height=0.52\linewidth]{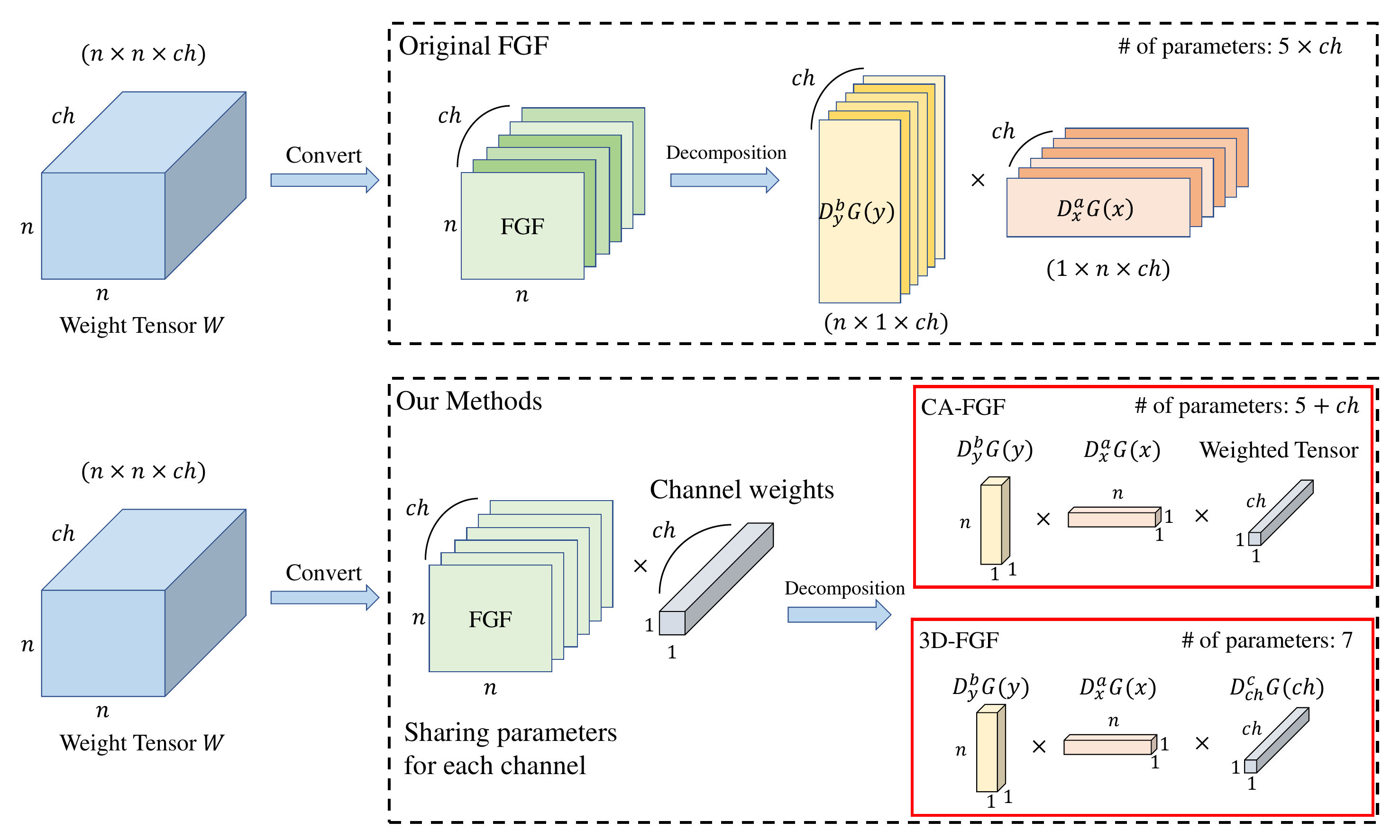}
    \caption{
    The details of FGF Transformation in the FGFS. The composition of the original FGF for a single kernel, where independent FGFs are applied to each channel, reduces the number of parameters from $n \times n \times ch$ to $5 \times ch$. On the other hand, the FGFS contains two FGF forms, the CA-FGF and the 3D-FGF. The FGFS shares the parameters of the FGF across all channels. The CA-FGF can reduce the parameter count to $5 + ch$ with the channel-attention mechanism. Additionally, to achieve further compression, the 3D-FGF constraints the channel weights and reduces the parameter count to just $7$.}
    \label{fig:cafgf}
\end{figure*}

As shown in Fig.~\ref{fig:cafgf}, we share the five parameters for all channels to achieve the maximum compression ratio. However, sharing these parameters would cause the model accuracy degradation since all FGFs for each channel are the same. To address this, we would introduce the weights for each channel in our CA-FGF. The mechanism of the weighted channel can emphasize the essential features and suppress the insignificant ones to improve the model's accuracy. Furthermore, the mechanism of the weighted channel in the CA-FGF can reduce the parameters from $5\times ch$ to $5+ch$, where $ch$ is the channel number of the FGF. 
In the 3D-FGF, we utilize the fractional derivatives of the Gaussian function to constrain the weights of channels to improve model generalizability. Therefore, the 3D-FGF can be defined as follows: 
\begin{equation} \label{eq:3dfgf}
\begin{aligned}
F_{3d} &= D^a_x D^b_y D^c_{ch} G(x,y,ch) = D^a_x D^b_y D^c_{ch} G(x)G(y)G(ch) \\
       &= D^a_x G(x) \times D^b_y G(y) \times D^c_{ch} G(ch),
\end{aligned}
\end{equation}
where $c$ is the order of the fractional derivatives in the channel direction. By using the fractional derivatives of the Gaussian function as the weights of channels, the 3D-FGF can achieve the maximum compression ratio. The trainable parameters can be reduced from $5\times ch$ to seven, including the fractional order ($a, b, c$), the offset of the center($x_0, y_0, ch_0$), and the standard deviation of the Gaussian distribution ($\sigma$). Same as the original FGF, the ranges of parameters are defined as follows: $a \in [0,2]$, $b \in [0,2]$, $c \in [0,2]$, $x_0 \in (-\infty,\infty)$, $y_0 \in (-\infty, \infty)$, $ch_0 \in (-\infty,\infty)$, and $\sigma \in (0,\infty)$.

\subsection{Adaptive Unstructured Pruning (AUP)}

Adaptive Unstructured Pruning (AUP) eliminates the redundant parameters by removing those with absolute values smaller than the pruning threshold during each round. The pruning threshold is determined by the predefined percentage $p_{r}$ of non-zero parameters to be removed in each round. However, removing the numerous parameters in the model causes a decrease in its performance. To prevent a significant drop in accuracy after pruning, the model is fine-tuned after each round of pruning to recover its accuracy. Suppose the fine-tuning model after a pruning round cannot reach the accuracy threshold $\theta_{acc}$. In that case, this pruning round will be abandoned, and the model will reload the previous round's model for additional fine-tuning before continuing pruning. When the model can not reach the accuracy threshold $\theta_{acc}$ after several pruning rounds, we will reduce the predefined percentage $p_{r}$ and continue attempting to prune. Once the sparsity ratio reaches the target, we fine-tune the model to restore its accuracy to a higher level.

\subsection{Fractional Gaussian Filter and Pruning (FGFP)}
The Fig.~\ref{fig:Overall_HFGFS_Method} demonstrates the fractional Gaussian filter and pruning (FGFP) framework. First, we convert the filters of the pre-trained model into the FGF. Previous studies \cite{Zamora_Convolutional, Llanza_Deep} have demonstrated that applying fractional Gaussian filters (FGF) in deeper layers with larger input channels yields higher compression ratios with minimal accuracy degradation. Hence, we apply FGF to the deeper layers with larger input channels. After replacing the large layer with the FGF, we apply adaptive unstructured pruning (AUP) methods to the remaining layers. In this step, we define a layer-wise pruning ratio $p_{r}$, typically between 3\% and 6\%, to control the proportion of parameters removed in each pruning round. Moreover, we independently perform the adaptive unstructured pruning for each layer to prevent concentrating the pruned parameters in a single layer due to the collectively smaller parameter values. After all layers are processed with the FGF transform and AUP, we can obtain the compressed model with the FGFP framework.
\section{Experimental Results}

\begin{table*}[ht]
  \centering
  \caption{Results for ResNet-20, ResNet-32 and WRN-28-10 on CIFAR-10 dataset. “$\ast$” denotes validation accuracy since the corresponding work does not provide test accuracy.}
  \label{tab:CIFAR10}
  \begin{tabular}{@{}lccccc@{}}
    \toprule[2pt]
    \multirow{2}{*}[-0.6ex]{Method} & \multirow{2}{*}[-0.6ex]{Post-Trained Model Type} & \multicolumn{3}{c}{Top-1 Accuracy (\%)} & \multirow{2}{*}[-1.0ex]{Parameter CR (\%)}\\
    \cmidrule(lr){3-5}
    && Baseline & Compressed & $\Delta \downarrow$ &\\
    \midrule[0.3pt]
    \multicolumn{6}{c}{ResNet-20} \\
    \midrule[0.3pt]
    SCOP \cite{Tang_SCOP}                & Sparse                     & 92.22 & 90.75  & 1.47  & 56.3\\
    Hinge \cite{Li_Hinge}                & Low-Rank + Sparse          & 92.54 & 91.84  & 0.70  & 55.5\\
    \rowcolor{gray!20}
    FGFP(CA-FGF) (ours)           & Fractional Filter + Sparse & 91.34 & 90.77  & \textbf{0.57}  & 59.3\\
    \rowcolor{gray!20}
    FGFP(3D-FGF) (ours)           & Fractional Filter + Sparse & 91.34 & 90.34  & 1.00  & \textbf{66.7}\\
    \hdashline
    PSTRN-M \cite{Li_Heuristic}          & Low-Rank                   & 91.25 & 89.30  & 1.95  & 85.2\\
    ELRT \cite{Sui_ELRT}                 & Low-Rank                   & 91.25 & 89.64  & 1.61  & 83.4\\
    TDLC \cite{Liu_TDLC}                 & Low-Rank                   & 91.25 & 88.58  & 2.65  & 80.5\\
    \rowcolor{gray!20}
    FGFP(CA-FGF) (ours)           & Fractional Filter + Sparse & 91.34 & 90.20  & \textbf{1.14}  & 81.5\\
    \rowcolor{gray!20}
    FGFP(3D-FGF) (ours)        & Fractional Filter + Sparse & 91.34 & 89.82  & 1.52  & \textbf{85.2}\\
    \midrule[0.3pt]
    \multicolumn{6}{c}{ResNet-32} \\
    \midrule[0.3pt]
    SCOP \cite{Tang_SCOP}                & Sparse                     & 92.66 & 92.13 & 0.53 & 56.2\\
    PSTRN-S \cite{Li_Heuristic}          & Low-Rank                   & 92.49 & 91.43 & 1.06 & 60.9\\
    \rowcolor{gray!20}    
    FGFP(CA-FGF) (ours)  & Fractional Filter + Sparse & 92.64 & 92.11 & \textbf{0.53} & \textbf{76.1}\\
    \rowcolor{gray!20}
    FGFP(3D-FGF) (ours)        & Fractional Filter + Sparse & 92.64 & 91.92 & 0.72 & \textbf{76.1}\\
    \hdashline
    PSTRN-M \cite{Li_Heuristic}          & Low-Rank                   & 92.49 & 90.59 & 1.90 & 80.4\\
    ELRT \cite{Sui_ELRT}                 & Low-Rank                   & 92.49 & 91.21 & 1.28 & 80.4\\
    \rowcolor{gray!20}
    FGFP(CA-FGF) (ours)   & Fractional Filter + Sparse & 92.64 & 91.85 & \textbf{0.79} & \textbf{80.4}\\
    \rowcolor{gray!20}
    FGFP(3D-FGF) (ours)  & Fractional Filter + Sparse & 92.64 & 91.80 & 0.84 & \textbf{80.4}\\
    \midrule[0.3pt]
    \multicolumn{6}{c}{WRN-28-10} \\
    \midrule[0.3pt]
    GrowEfficient \cite{Yuan_Growing}    & Sparse                     & 96.20 & 95.30 & 0.90* & 90.7\\
    BackSparse \cite{Zhou_Efficient}     & Sparse                     & 96.20 & 95.60 & 0.60* & 91.6\\
    \rowcolor{gray!20}
    FGFP(CA-FGF) (ours) & Fractional Filter + Sparse & 94.78 & 93.68 & 1.10* & 91.6\\
    \rowcolor{gray!20}
    FGFP(3D-FGF) (ours)           & Fractional Filter + Sparse & 94.78 & 94.24 & \textbf{0.54*} & \textbf{96.8}\\
    \bottomrule[2pt]
  \end{tabular}

\end{table*}
\begin{table*}[ht]
  \centering
  \caption{Results for ResNet-18 and ResNet-50 on ImageNet2012 dataset.}
  \begin{tabular}{@{}lccccc@{}}
    \toprule[2pt]
    \multirow{2}{*}[-0.6ex]{Method} & \multirow{2}{*}[-0.6ex]{Post-Trained Model Type} & \multicolumn{3}{c}{Top-1 Accuracy (\%)} & \multirow{2}{*}[-1.0ex]{Parameter CR (\%)}\\
    \cmidrule(lr){3-5}
    && Baseline & Compressed & $\Delta \downarrow$ &\\
    \midrule[0.3pt]
    \multicolumn{6}{c}{ResNet-18} \\
    \midrule[0.3pt]
    FR \cite{Chu_Low-rank}               & Low-Rank                   & 69.76 & 69.04 & 0.72 & 57.0\\
    LRPET \cite{Guo_Compact}             & Low-Rank                   & 69.76 & 67.87 & 1.89 & 50.3\\
    \rowcolor{gray!20}
    FGFP(3D-FGF) (ours)      & Fractional Filter + Sparse & 69.30 & 68.61 & \textbf{0.69} & 60.1\\
    \rowcolor{gray!20}
    FGFP(3D-FGF) (ours)      & Fractional Filter + Sparse & 69.30 & 68.28 & 1.02 & \textbf{74.7}\\
    \midrule[0.3pt]
    \multicolumn{6}{c}{ResNet-50} \\
    \midrule[0.3pt]
    EDP \cite{Ruan_EDP}                  & Low-Rank + Sparse          & 75.90 & 75.34 & 0.56 & 43.9\\
    ARPruning \cite{Yuan_ARPruning}      & Sparse                     & 76.15 & 72.31 & 3.84 & 56.8\\
    SFI-FP \cite{Liu_Soft}               & Sparse                     & 76.15 & 75.21 & 0.94 & 57.3\\
    CORING \cite{Van_Efficient}          & Sparse                     & 76.15 & 75.55 & 0.60 & 56.7\\
    \rowcolor{gray!20}
    FGFP(3D-FGF) (ours)           & Fractional Filter + Sparse & 76.16 & 75.64 & \textbf{0.52} & \textbf{57.4}\\
    \hdashline
    Stable \cite{Phan_Stable}            & Low-Rank                   & 76.15 & 74.68 & 1.47 & 60.2\\
    CC \cite{Li_Towards}                 & Low-Rank + Sparse          & 76.15 & 74.54 & 1.61 & 58.6\\
    \rowcolor{gray!20}
    FGFP(3D-FGF) (ours)           & Fractional Filter + Sparse & 76.16 & 75.42 & \textbf{0.74} & \textbf{62.7}\\
    \hdashline
    AHC-A \cite{Wang_All-in-One}         & Sparse                     & 76.20 & 74.70 & 1.50 & 63.4\\
    LRPET-S \cite{Guo_Compact}           & Low-Rank                   & 76.15 & 73.72 & 2.43 & 64.0\\
    \rowcolor{gray!20}
    FGFP(3D-FGF) (ours)          & Fractional Filter + Sparse & 76.16 & 74.82 & \textbf{1.34} & \textbf{66.8}\\
    \hdashline
    NORTON \cite{Pham_Enhanced}          & Low-Rank + Sparse          & 76.15 & 74.00 & 2.15 & 68.8\\
    \rowcolor{gray!20}
    FGFP(3D-FGF) (ours)          & Fractional Filter + Sparse & 76.16 & 74.53 & \textbf{1.63} & \textbf{69.1}\\
    \bottomrule[2pt]
  \end{tabular}
  \label{tab:ILSVRC2012}
\end{table*}
\subsection{Experimental Settings}

\textbf{Datasets.}
To evaluate the performance of our proposed method, we employed two widely used benchmarks in image classification: CIFAR-10 and ImageNet2012. CIFAR-10 is a classic small-scale image dataset consisting of 10 classes, with 50K training images and 10K test images, each of size $32 \times 32$. We further partitioned the CIFAR-10 training set during training into a training subset of 45K images and a validation subset of 5K images. ImageNet2012, on the other hand, is a large-scale dataset for image classification, comprising approximately 1.28M training images, 50K validation images, and 100K test images.

\textbf{Networks.}
On the CIFAR-10 dataset, we evaluated the FGFP with CA-FGF and 3D-FGF using ResNet-20, ResNet-32 \cite{ResNet}, and WRN-28-10 \cite{WRN}. In addition, since ResNet-18 and ResNet-50 \cite{ResNet} are larger networks, we apply the FGFP with the 3D-FGF in ResNet-18 and ResNet-50 for a higher compression ratio on the ImageNet2012 dataset.

\textbf{Evaluation Metrics.}
The model is evaluated using the accuracy drop and the number of parameters required. Considering the model's performance, top-1 accuracy is utilized on classification tasks. Also, the parameter compression ratio (CR) is defined as the percentage reduction in the number of parameters compared to the original model.

\textbf{Configurations.}
 All experiments use the stochastic gradient descent (SGD) optimizer. The batch sizes are 128 and 256 for CIFAR-10 and ImageNet2012, respectively. The learning rates in the FGF training stage are 0.1 for all model architectures. The learning rates in the adaptive unstructured pruning stage are set at 0.01 to 0.001 for ResNet-20 and ResNet-32, and 0.0001 for WRN-28-10, ResNet-18, and ResNet-50, respectively.

\subsection{Results and Analysis}
\textbf{CIFAR-10.}
Table~\ref{tab:CIFAR10} presents the comparison results between our FGFP and recent works on ResNet-20, ResNet-32, and WRN-28-10. When we evaluate the performance of ResNet-20 on the test set, the compression ratio of FGFP can reach higher than 80\% no matter using both the 3D-FGF and the CA-FGF. Meanwhile, the reduction of accuracy is less than the PSTRN-M \cite{Li_Heuristic}, the ELRT \cite{Sui_ELRT}, and the TDLC \cite{Liu_TDLC}. Similarly, on ResNet-32, FGFS also achieves excellent performance, reaching a compression ratio of 80.4\% while maintaining higher accuracy than both PSTRN-M and ELRT. For WRN-28-10, the FGFP, especially with the 3D-FGF, can compress the network by 96.8\% and significantly outperform the GrowEfficient~\cite{Yuan_Growing} and BackSparse~\cite{Zhou_Efficient}.

\textbf{ImageNet2012.}
To evaluate the scalability of the FGFP, we perform experiments on the ImageNet2012 dataset using the ResNet-18 and ResNet-50 architectures, which are shown in Table~\ref{tab:ILSVRC2012}. On ResNet-18, FGFP with 3D-FGF achieves a 74.7\% compression ratio with only about 1\% accuracy degradation, demonstrating superior performance compared to recent works such as FR \cite{Chu_Low-rank} and LRPET \cite{Guo_Compact}. Similarly, on ResNet-50, the FGFP with the 3D-FGF also achieves a higher compression ratio, 69.1\%, than other compression methods, while the accuracy of the FGFP only drops 1.63\%. Also, the FGFP can maintain better performance when the network model is compressed by various compression ratios.

\subsection{Ablation Study}
\textbf{Comparison of FGFP and AUP.}
We conducted an ablation study on CIFAR-10 using ResNet-20 to demonstrate that combining fractional filters and pruning outperforms using adaptive unstructured pruning alone. The experimental results are shown in Table~\ref{tab:compress_vs_prune}. We aim to compress the network model to the specified number of parameters with adaptive unstructured pruning (AUP) and the FGFP, which combines the AUP and the CA-FGF. In Table~\ref{tab:compress_vs_prune}, the FGFP shows superior performance compared to using AUP alone, achieving a 0.33\% accuracy improvement when the remaining parameters are reduced to 0.07M.

\begin{table}[htb]
  \centering
  \caption{Performance comparison between FGFP and AUP for ResNet-20 on CIFAR-10.}
  \begin{tabular}{@{}lcc@{}}
    \toprule[2pt]
    Method & $\Delta$Acc. $\downarrow$(\%) & \#Parameter\\
    \midrule
    AUP                 &         1.19  & 0.07M\\
    FGFP(CA-FGF + AUP) & \textbf{0.86} & 0.07M\\
    \bottomrule[2pt]
  \end{tabular}
  \label{tab:compress_vs_prune}
\end{table}


\textbf{Comparison of FGFP(CA-FGF) and FGFP(3D-FGF).}
To analyze the relationship between CA-FGF, 3D-FGF, and adaptive unstructured pruning (AUP), we converted the 28th, 29th, 30th, and 31st layers of ResNet-32 into FGF layers while performing AUP on the remaining layers. As shown in Table \ref{tab:2D_vs_3D_UPFF}, when only FGF conversion was applied, the accuracy of CA-FGF was 0.25\% higher than that of 3D-FGF. After applying AUP, when the total number of parameters was reduced to 0.07M, the accuracy of CA-FGF remained 0.16\% higher than that of 3D-FGF. However, as parameters dropped to 0.05M, excessive AUP pruning caused significant feature loss, resulting in an additional 0.31\% accuracy drop in CA-FGF compared with 3D-FGF.
\begin{table}[htb]
  \centering
  \caption{Comparison of accuracy and compression ratio between FGFP using CA-FGF and 3D-FGF for ResNet-32 on CIFAR-10.}
  \begin{tabular}{@{}lcc@{}}
    \toprule[2pt]
    Method & $\Delta$Acc. $\downarrow$(\%) & \#Parameter\\
    \midrule[0.3pt]
    CA-FGF        & \textbf{0.36} & 0.33M\\
    3D-FGF        &         0.61  & 0.32M\\
    \hdashline
    FGFP(CA-FGF) & \textbf{1.11} & 0.07M\\
    FGFP(3D-FGF) &         1.27  & 0.07M\\
    \hdashline
    FGFP(CA-FGF) &         2.26  & 0.05M\\
    FGFP(3D-FGF) & \textbf{1.95} & 0.05M\\
    \bottomrule[2pt]
  \end{tabular}
  \label{tab:2D_vs_3D_UPFF}
\end{table}
\section{Conclusion}
In this paper, we present the novel fractional Gaussian filter and Pruning (FGFP) framework for network model compression. There are two principal mechanisms in the FGFP: the fractional Gaussian filter (FGF) and the adaptive unstructured pruning (AUP).
We integrate fractional-order differential calculus with the Gaussian function to construct the FGF, incorporating the Gr\"unwald–Letnikov fractional derivatives for simplification. By sharing the parameters of the FGF and utilizing the technique of channel attention, the number of parameters in the network can be reduced to seven. Moreover, we apply the AUP to our FGFP to achieve the maximum compression ratio and maintain the accuracy of the models. We also utilize comprehensive experiments on CIFAR-10 and ImageNet2012 to validate the effectiveness of our proposed method. According to the experimental results, the 85.2\% compression ratio can be achieved with only a 1.52\% degradation in accuracy by using the FGFP to compress the ResNet-20. Also, the FGFP can achieve the compression ratio of 69.1\% while the accuracy only decreases by 1.63\% on ResNet-50 with the ImageNet2012 dataset. In summary, the FGFP, which combines the fractional Gaussian filter and adaptive unstructured pruning, is a promising solution to mitigate parameter redundancy in modern deep neural networks and achieves substantial model compression with minimal accuracy degradation.


\bibliography{main}
\bibliographystyle{icml2025}

\end{document}